\documentclass[conference]{IEEEtran}
\IEEEoverridecommandlockouts
\usepackage{cite}
\usepackage{amsmath,amssymb,amsfonts}
\usepackage{algorithmic}
\usepackage[linesnumbered]{algorithm2e}
\usepackage{graphicx}
\usepackage[marginal]{footmisc}

\usepackage{flushend,cuted}
\usepackage{subfigure}
\usepackage{multirow}
\usepackage{textcomp}
\usepackage{xcolor}
\def\BibTeX{{\rm B\kern-.05em{\sc i\kern-.025em b}\kern-.08em
    T\kern-.1667em\lower.7ex\hbox{E}\kern-.125emX}}
\begin{document}

\title{DualHGNN: A Dual Hypergraph Neural Network for Semi-Supervised Node Classification based on Multi-View Learning and Density Awareness
}

\author{
\IEEEauthorblockN{Jianpeng Liao}
\IEEEauthorblockA{
\textit{South China University of Technology} \\
Guangzhou, Guangdong, China\\
sejianpengliao@mail.scut.edu.cn}
\and
\IEEEauthorblockN{Jun Yan}
\IEEEauthorblockA{
\textit{Concordia University} \\
Montreal, Quebec, Canada\\
jun.yan@concordia.ca}
\and
\IEEEauthorblockN{Qian Tao\IEEEauthorrefmark{1}}
\IEEEcompsocitemizethanks{\IEEEauthorrefmark{1}Corresponding author.}
\IEEEauthorblockA{
\textit{South China University of Technology}\\
Guangzhou, Guangdong, China\\
\textit{Pazhou Lab}, Guangzhou, Guangdong, China\\
taoqian@scut.edu.cn}
}
\maketitle
\footnote{This work has been accepted by 2023 International Joint Conference on Neural Networks (IJCNN 2023).}

\begin{abstract}
  Graph-based semi-supervised node classification has been shown to become a state-of-the-art approach in many applications with high research value and significance.
  Most existing methods are only based on the original intrinsic or artificially established graph structure which may not accurately reflect the ``true” correlation among data and are not optimal for semi-supervised node classification in the downstream graph neural networks.
  Besides, while existing graph-based methods mostly utilize the explicit graph structure, some implicit information, for example, the density information, can also provide latent information that can be further exploited.
  To address these limitations, this paper proposes the Dual Hypergraph Neural Network (DualHGNN), a new dual connection model integrating both hypergraph structure learning and hypergraph representation learning simultaneously in a unified architecture. 
  The DualHGNN first leverages a multi-view hypergraph learning network to explore the optimal hypergraph structure from multiple views, constrained by a consistency loss proposed to improve its generalization.
  Then, DualHGNN employs a density-aware hypergraph attention network to explore the high-order semantic correlation among data points based on the density-aware attention mechanism.
  Extensive experiments are conducted in various benchmark datasets, and the results demonstrate the effectiveness of the proposed approach.
\end{abstract}


\section{Introduction}
Over the last few years, Graph Neural Networks (GNNs) have attracted much attention because of their ability to effectively deal with graph-structured data and achieve amazing performance, and have been  widely used for many machine learning tasks including computer vision \cite{DBLP:conf/cvpr/LiLCCYY20}, recommendation systems \cite{DBLP:conf/icml/YuQ20}, neural machine translation \cite{DBLP:conf/acl/YinMSZYZL20}, and others. Compared with the traditional neural networks that encode every single data separately, GNNs can encode the graph structure of different input data through a graph message propagation mechanism, which allows it to obtain more information than the single data encoding for neural networks. 

Graph-based semi-supervised learning, which can exploit the connectivity relationship between small amounts of labeled samples and a relatively large number of unlabeled samples to improve the performance of deep neural networks, has been shown to be one of the most effective approaches for semi-supervised node classification. Graph-based semi-supervised node classification has seen applications in various fields, such as predicting the customer type of users in e-commerce \cite{DBLP:journals/pvldb/EswaranGFMK17}, assigning scientific papers from a citation network into topics \cite{GLCN,GCNs}, and credit card fraud detection \cite{DBLP:conf/kdd/Jin0LTWT20}. %This also makes graph-based semi-supervised node classification have high research value and important significance.

To date, a large number of graph-based semi-supervised node classification methods have been proposed \cite{GCNs, GATs, GraphSAGE}. Most of these methods focused only on the pairwise connections among data. However, the data correlation in real practice could be beyond pairwise relationships and even more complicated. Under such circumstances, only exploring the pairwise connections and modeling it as a graph may lose the high-order semantic correlation among data. The traditional structure with simple graphs cannot fully formulate the data correlation and thus limits the application of GNN models \cite{HGNN}. To tackle this challenge, hypergraph neural networks (HGNNs) have been proposed, introducing hyperedges that can link any number of nodes to improve the learning performance. Compared with the simple graph, the hyperedges in HGNNs allows the latter to more effectively represent the high-order semantic relationship among data \cite{HGNN, HGATs, DHGNN}. This work will also leverage the hypergraph to explore the high-order semantic correlation among data for semi-supervised node classification.

In this study, we will mainly focus on two challenges in graph-based semi-supervised node classification.
First, we noted that much of the success of graph and hypergraph neural networks is attributed to the graph-structure data offered to them.
In general, the data we provide to GNNs either have a known intrinsic graph structure, such as citation networks or are a human-established graph that we construct for it, such as $k$-nearest neighbor graph. 
However, we cannot guarantee that the original intrinsic or artificially established graph is optimal for semi-supervised node classification in the downstream graph neural networks.
Besides, the original graph is usually constructed from the original feature space in which the similarity between samples may not be accurately measured. In other words, the original graph may have some redundant or missing edges, and thus it may not accurately reflect the ``true" correlation among data.
What’s more, the human-established $k$-nearest neighbor graph is mainly based on a fixed and single similarity measurement function, which may not be suitable for accurately measuring the similarity between all samples. 
Accordingly, this calls for accurate modeling and learning techniques to obtain a suitable graph and hypergraph structure.

Second, most existing graph-based semi-supervised node classification methods mostly only utilized the explicit graph structure information \cite{GCNs, GATs, HGNN}. One of the most challenging problems for semi-supervised learning is how to exploit the implicit information among data to improve model performance. Some implicit information among data, for example, the density information, has been demonstrated to provide important clues for semi-supervised node classification \cite{DBLP:conf/cvpr/LiLCCYY20}, yet it is rarely exploited in depth. Li \emph{et al.} \cite{DBLP:conf/cvpr/LiLCCYY20} first exploited density information for graph-based deep semi-supervised visual recognition. Yet it is also only based on the exploration of the graph-structure relationships among data, while high-order semantic correlation has been ignored. Inspired by this, we decided to explore density information among data on hypergraph structure to improve semi-supervised node classification accuracy in this work.

To tackle these two challenges, we propose the Dual Hypergraph Neural Network (DualHGNN), a dual connection model containing two sub-networks that perform hypergraph structure learning and hypergraph representation learning for graph-based semi-supervised node classification.
For the first challenge, DualHGNN adopts a multi-view hypergraph learning network to learn the hypergraph structure from multiple views. By adopting different learnable similarity measure functions on each view, we can measure the sample similarity more accurately. By introducing a consistency loss, DualHGNN can effectively improve the generalization ability of hypergraph learning.
For the second, DualHGNN employs a density-aware hypergraph attention network to exploit density information on hypergraph explicitly to improve semi-supervised node classification performance. We define a density rule for hypergraph and structure a density-aware attention mechanism. Based on density-aware attention, DualHGNN can effectively improve hypergraph representation learning.
In short, DualHGNN jointly optimizes the multi-view hypergraph learning network and the density-aware hypergraph attention network to learn the optimal hypergraph suitable for downstream graph-based semi-supervised node classification tasks. Meanwhile, based on the suitable hypergraph, we can improve the performance of the density-aware hypergraph attention network. As shown in the experiments, the combination of two HGNNs effectively allows the proposed architecture to achieve higher classification performance.

The main contributions of can be summarized as follows.
\begin{itemize}
\item A novel Dual Hypergraph Neural Network (DualHGNN) is proposed, integrating both hypergraph structure learning and hypergraph representation learning simultaneously in a unified network architecture for semi-supervised node classification. 
\item A new multi-view hypergraph learning network is proposed to learn an optimal hypergraph suitable for downstream semi-supervised node classification from multiple views with different learned similarity measure functions, constrained by a consistency loss to improve its generalization ability. 
\item The explicit density information of hypergraphs is leveraged to propose a density-aware hypergraph attention network. A density rule for hypergraphs is defined, and a density-aware attention mechanism is developed to effectively improve the performance of hypergraph representation learning.
\item Extensive experiments have been conducted to demonstrate the effectiveness of the DualHGNN for semi-supervised node classification. The ablation study further proved the validity of the multi-view hypergraph learning and the density-aware attention mechanism.
\end{itemize}

\section{Related Work}
\label{RelatedWork}

\subsection{Graph Neural Networks}  
The core idea of graph neural networks (GNNs) is graph message propagation \cite{DBLP:journals/tnn/WuPCLZY21}, which can be divided into spectral-based approaches \cite{GCNs} and spatial-based approaches \cite{GraphSAGE, GATs}. 
Graph convolutional networks (GCNs) \cite{GCNs} performed label prediction based on graph neighborhood aggregation, which provided a novel idea for graph spectral information propagation. 
By adopting a self-attention layer, Velickovic \emph{et al.} proposed graph attention networks (GATs) \cite{GATs} to perform attention neighborhood aggregation.
Wu \emph{et al.} \cite{SGC} removed the nonlinear activation function and collapsed weight matrices from GCNs \cite{GCNs} and proposed a simplifying GNNs.
Liu \emph{et al.} \cite{ElasticGNN} proposed ElasticGNN by introducing $L1$ and $L2$ regularization and providing an elastic message passing scheme to enhance the local smoothness of the graph.
Recently, Duan \emph{et al.} \cite{IJCNN:DCSGCN} proposed a dual cost-sensitive graph convolutional network (DCSGCN) to tackle the imbalanced graph learning problem.
However, the simple graph structure may not fully formulate the high-order data correlation, for which hypergraphs can provide an effective solution. %toward better semi-supervised node classification.

\subsection{Hypergraph Neural Networks} 
A hypergraph is a generalization of graphs to model the high-order semantic correlation among data. 
Shi \emph{et al.} \cite{DBLP:journals/tnn/ShiZZMZGS19} adopted a hypergraph learning process to optimize the high-order correlation among data. 
Feng \emph{et al.} \cite{HGNN} proposed a hypergraph neural network (HGNN)  to perform the node-edge-node transform through hyperedge convolution operations.
Hypergraph attention networks (HGATs) \cite{HGATs} introduced the attention mechanism into hypergraph neural networks to encode the high-order data correlation.
Jiang \emph{et al.} \cite{DHGNN} integrated dynamic hypergraph construction and hypergraph convolution modules to propose dynamic hypergraph neural networks (DHGNN) further improve hypergraph representation learning.
Recently, many improved methods have been proposed, including hypergraph label propagation networks (HLPN) \cite{DBLP:conf/aaai/ZhangWCZWZ020} and hypergraph convolution and hypergraph attention (HCHA) \cite{DBLP:journals/pr/BaiZT21}, among others.  
However, most of these studies focused only on hypergraph representation learning based on the original hypergraph that may not accurately reflect the “true” data correlation and are not optimal for the downstream HGNNs, and this motivates the accurate learning of a suitable hypergraph structure to improve the performance of HGNNs.

\subsection{Graph-based Semi-supervised Node Classification} 
Graph-based semi-supervised learning methods are one of the most effective approaches for semi-supervised node classification. 
Hamilton \emph{et al.} \cite{GraphSAGE} proposed an inductive graph neural network GraphSAGE extending graph data processing to large graphs.
Gasteiger \emph{et al.} \cite{APPNP} proposed APPNP by introducing a personalized PageRank propagation scheme, achieving graph information propagation in a larger neighborhood.
Yet they all neglected the learning of graph structure.
Jiang \emph{et al.} \cite{GLCN} introduced a graph learning module to learn an optimal graph structure that makes GCNs \cite{GCNs} better for semi-supervised learning. In the same way, we introduce a hypergraph learning module in our method.
Rong \emph{et al.} \cite{DropEdge} randomly removed a certain number of edges from the input graph to realize data enhancement.
Similarly, Tang \emph{et al.} \cite{GRAND} proposed GRAND by designing a random propagation strategy based on the drop node mechanism.
Yet both \cite{DropEdge} and \cite{GRAND} only use a sub-optimal graph structure for semi-supervised node classification.
Song \emph{et al.} \cite{KDD:OAGS} formulated a Bayesian probabilistic model, obtained the posterior distribution from the downstream classification module, and employed a variational inference method to an optimal graph.
Lee \emph{et al.} \cite{SIGIR:GraFN} proposed GraFN to learn discriminative node representations through supervised and unsupervised consistency between two augmented graphs.
Li \emph{et al.} \cite{IJCNN:CoGNet} proposed a cooperative dual-view graph neural network regarding different views as the reasoning processes of two GNN models. Unlike \cite{IJCNN:CoGNet}, we adopt a dual connection model in our DualHGNN, which is based on the effective combination of two hypergraph neural networks.
In addition, most of these methods only utilized the explicit graph structure information, calling for an effective mechanism to explore implicit information in the hypergraphs, such as the density, to improve the performance of semi-supervised node classification.

\section{The DualHGNN Architecture}
\label{Methods}

\begin{figure*}[h!]
\centering
\includegraphics[width=\linewidth]{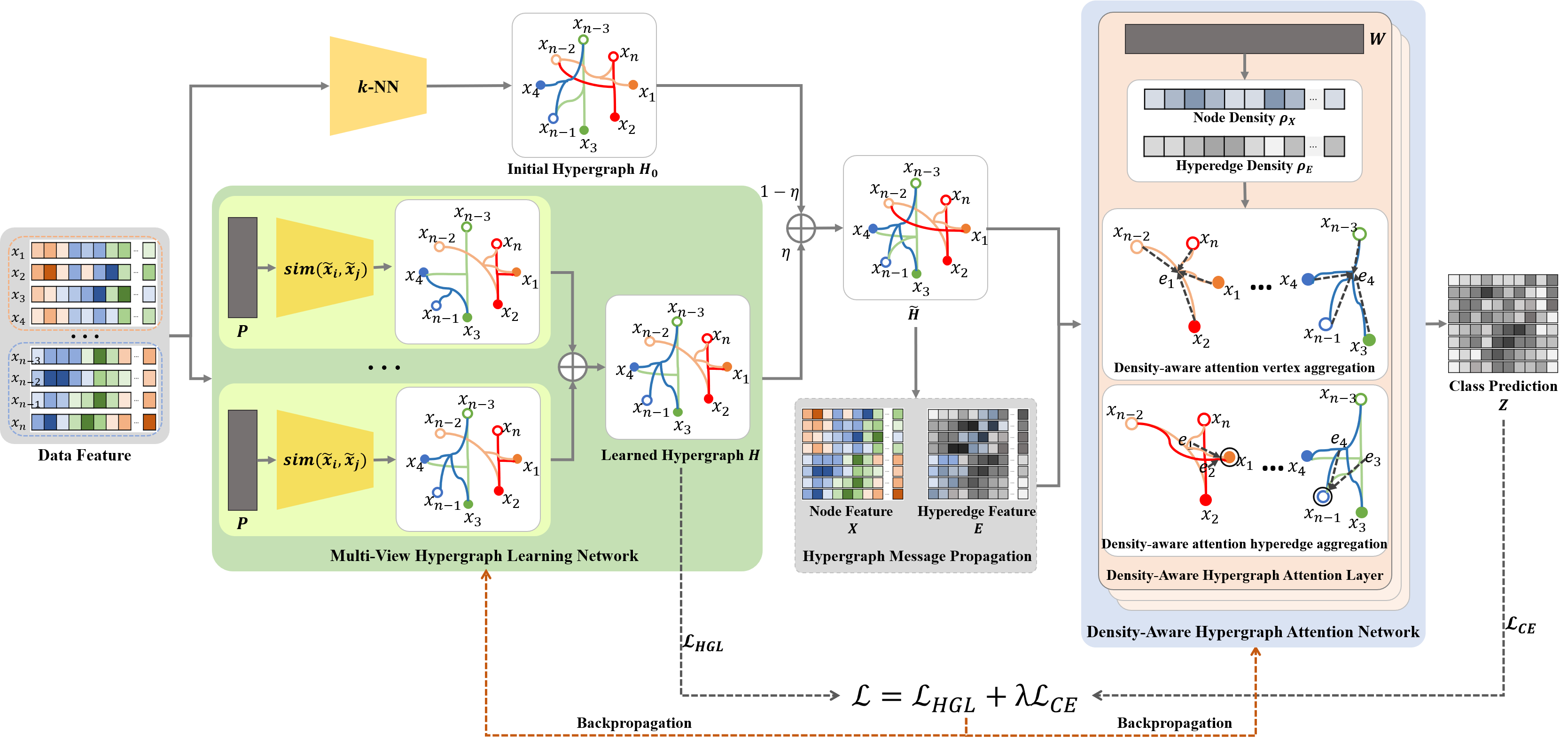}
\caption{Overview of the proposed DualHGNN framework.}
\label{DualHGNN}
\end{figure*}

The proposed DualHGNN is shown in Figure~\ref{DualHGNN}. 
The DualHGNN first adopts a multi-view hypergraph learning network to learn a suitable hypergraph structure on multi-view with different similarity measure functions and outputs a new hypergraph.
Subsequently, the DualHGNN employs a density-aware hypergraph attention network based on a density-aware attention mechanism to perform hypergraph representation learning for class prediction.
We linearly combine the losses calculated from the output of two sub-networks and perform backpropagation to update the parameters of these two modules at the same time.
The specific designs are elaborated as follows.

\subsection{Multi-View Hypergraph Learning Network}

A hypergraph can be formulated as $\mathcal{G}=(\mathcal{V}, \mathcal{E})$, which includes a set of vertex $\mathcal{V}$ and a set of hyperedges $\mathcal{E}$. Let $X=[x_{1},x_{2},\dots,x_{n}]\in \mathbb{R}^{n \times d} $ be the collection of $n$ data vectors of $d$ dimensions, where $x_{i}$ denotes the feature vector of $i$-th sample. The structure of hypergraph can be denoted by an incidence matrix $H\in \mathbb{R}^{n \times m}$, in which $H_{x_{i},e_{k}}=1$ indicates that the node $x_{i}$ is connected by the hyperedge $e_{k}$, otherwise $H_{x_{i},e_{k}}=0$, and $n$ and $m$ are the numbers of nodes and hyperedges, respectively, in the hypergraph. 
% The labels of node can be represented as $Y=[y_{1},y_{2},\dots,y_{n}]\in \mathbb{R}^{n \times c}$ and each $y_{i}\in \mathbb{R}^{c}$ corresponds to a $c$-dimensional one-hot label representation where $c$ is the number of classes.

The main idea of hypergraph learning is to learn an optimal hypergraph structure for semi-supervised node classification of the downstream hypergraph neural networks by jointly optimizing hypergraph structure learning and hypergraph representation learning.
In this paper, we proposed a multi-view hypergraph learning network to adaptively learn a suitable hypergraph.
The multi-view hypergraph learning network learns the hypergraph structure from multiple views with different learnable similarity measure functions to accurately fit the similarity between samples, which can avoid the defect that the single fixed distance measure function may not be able to accurately measure the similarity between all samples.
The final hypergraph structure can be obtained by merging the hypergraph learned in each view.

To avoid the influence of noise and data redundancy in the original feature space, we perform similarity learning in the low-dimensional embedding space.
We adopt a fully connected layer to map the feature matrix $X_{0}$ from the original feature space to the low-dimensional embedding space, which can be implemented by multiplying a learnable embedding matrix $P\in \mathbb{R}^{d \times p}$, that is,
\begin{equation}
    \tilde{X} = X_{0}P  .
\end{equation}

The similarity between samples can be measured by the function $\operatorname{sim}(\cdot )$ and obtained in a matrix $S$ as:
\begin{equation}
\label{eq2}
    S_{ij} = \operatorname{sim}(\tilde{x} _{i},\tilde{x} _{j}).
\end{equation}
To avoid the huge computational overhead brought by the fully connected graph, we perform sparse sampling for the similarity matrix $S$.
We employ a predefined threshold $\delta_{1}$ to filter out lower similarity values, which can be formulated as
\begin{equation}
\label{eq3}
\tilde{S}_{i j}=
\left\{\begin{array}{l}
\begin{matrix}
S_{i j}, & S_{i j} \geq \delta_{1} \\
0, & S_{i j}<\delta_{1}.
\end{matrix}
\end{array}\right.
\end{equation}
The hyperedges can then be constructed based on the sparse similarity matrix $\tilde{S}$ and obtain the learned hypergraph incidence matrix $H$.

Our multi-view hypergraph learning network can perform hypergraph learning on multi-view. Therefore, we can adopt different learnable similarity measure functions on each view, such as cosine similarity and inner product. 
The final hypergraph structure is obtained by the mean of hypergraph incidence matrices learned in each view, that is,
\begin{equation}
\label{eq5}
    H=\frac{1}{V} {\textstyle \sum_{v=1}^{V}} H^{(v)}, 
\end{equation}
where $V$ is the number of views, and $v$ represents the $v$-th view.

We introduce a consistency loss into the multi-view hypergraph learning network, aiming to constrain the similarity of hypergraph structures learned from each view. By introducing consistency loss, we can effectively use a large number of unlabeled samples to provide weak supervision to improve the generalization of the multi-view hypergraph learning network. 
The consistency loss can be defined as the sum of squared $L2$ distances between each output and its mean, that is,
\begin{equation}
    \mathcal{L}_{con} = \frac{1}{V}  {\textstyle \sum_{v=1}^{V}} {\left \| H^{(v)}- H \right \|}_2^{2}. 
\end{equation}

Considering that the original hypergraph may contain useful information. We merge the learned hypergraph with the original hypergraph and obtain the final incidence matrix $\tilde{H}$. It can be formulated as follow:
\begin{equation}
\label{eq8}
    \tilde{H}=\eta H+(1-\eta )H_{0} ,
\end{equation}
where $\eta$ is a trade-off parameter. $H_{0}$ can be a known intrinsic hypergraph or a human-established $k$-nearest neighbor hypergraph.

The loss function of the hypergraph learning network is
\begin{equation}
\label{eq9}
\mathcal{L}_{HGL} = \frac{\alpha}{n^{2}} \operatorname{tr}(\tilde{X} ^{\top} \hat{H} \tilde{X}) + 
\frac{\beta }{n^{2}} \left \| \tilde{H}  \right \| _{F}^{2} -
\frac{\gamma }{n} 1^{\top } \log{\hat{H} 1} +
\frac{\mu }{n^{2}} \mathcal{L}_{con},
\end{equation}
where $ \hat{H}=D_{v}^{-1/2} \tilde{H} D_{e}^{-1} \tilde{H}^{\top } D_{v}^{-1/2} $ is the hypergraph Laplacian, in which $D_{e}$ and $D_{v}$ are the diagonal matrices of the hyperedge degrees and the vertex degrees, respectively. $\alpha$, $\beta$, $\gamma$ and $\mu$ are hyperparameters. $\operatorname{tr}(\cdot)$ denotes the trace of matrix. $\left \| \cdot   \right \| _{F}$ is the Frobenius norm. $\cdot ^{\top }$ denotes transposition. This loss function contains four terms, where the first term restricts adjacent nodes to having similar features and learning a smooth incidence matrix, and the second one constrains learning a sparse hypergraph. The third term penalizes the formation of disconnected hypergraphs, and the last one is consistency loss.

\subsection{Density-Aware Hypergraph Attention Network}
The proposed DualHGNN uses the density information on the hypergraph structure to propose a density-aware hypergraph attention network. It integrates the density information with attention to structure a density-aware attention mechanism and performs hypergraph representation learning  through density-aware neighborhood feature aggregation, which is elaborated as follows.

The input of the density-aware hypergraph attention network includes the node feature matrix $X$ and the hyperedge feature matrix $E$. We perform hypergraph message propagation to obtain the node and the hyperedge feature matrix, that is,
\begin{equation}
\label{eq10}
     E = D_{e}^{-1/2} \tilde{H} ^{\top } D_{v}^{-1/2} X_{0},
\end{equation}
\begin{equation}
\label{eq11}
     X = D_{v}^{-1/2} \tilde{H} D_{e}^{-1/2} E.
\end{equation}

The density-aware hypergraph attention network mainly consists of two parts, density-aware attention vertex aggregation, and density-aware attention hyperedge aggregation. More detail about this will be described as follows.

\subsubsection{Density-aware attention vertex aggregation}
Density-aware attention vertex aggregation module integrates the density information of each node as a part of attention and then performs attention vertex aggregation to enhance hyperedge features. We define a density rule for each node, which can be defined as the sum of the similarities of neighbor nodes whose similarity with the target node is greater than a predefined threshold. The density of node $x_{i}$ can be formulated as
\begin{equation}
\label{eq12}
    \rho_{x_{i}}=\sum_{x_{k} \in \mathcal{N}_{x_{i}}}
    \left\{\begin{array}{l}
    \begin{matrix}
    \operatorname{sim}\left(x_{i}, x_{k}\right), & \text{ if } \operatorname{sim}\left(x_{i}, x_{k}\right) > \delta_{2} \\
    0, & \text{ if } 
    \operatorname{sim}\left(x_{i}, x_{k}\right) \le \delta_{2},
\end{matrix}
\end{array}\right.
\end{equation}
where $\mathcal{N}_{x_{i}}$ denotes the neighbors set of node $x_{i}$. $\delta_{2}$ is a predefined threshold. The similarity measure function $\operatorname{sim}(\cdot)$ can adopt cosine similarity in implementation. 

Intuitively, the higher the density of a node, the more neighbors that are similar to it. In other words, the target node is lying in a more densely distributed area. Based on the density-peak assumption \cite{rodriguez2014clustering}, the nodes with higher density are closer to the cluster center. Therefore, higher weights need to be assigned when performing neighborhood feature aggregation. While traditional attention mechanisms only consider feature similarity, which may be sub-optimal.
By fusing the density information, it can effectively avoid this defect and achieve more accurate attention neighborhood aggregation.

In the density-aware attention vertex aggregation module, we compute the attention weight of each node relative to the hyperedge it is on. 
We adopt an attention mechanism $\operatorname{Attention}(\cdot)$ to calculate the attention weight between node $x_{i}$ and hyperedge $e_{k}$, that is,
\begin{equation}
    a_{x_{i},e_{k}} = \operatorname{Attention}(Wx_{i}, We_{k}),
\end{equation}
where $W$ is the weight matrix that needs to be trained.

The density-aware attention mechanism can be then structured by combining the density information with the attention weight, which is shown in the following:
\begin{equation}
\label{eq14}
    \mathit{da}_{x_{i},e_{k}} = a_{x_{i}, e_{k}}+\tilde{\rho}_{x_{i}},
\end{equation}
where $\tilde{\rho}_{x_{i}} \in [0,\operatorname{max}(a_{X})]$ is the normalized density, and $a_{X}$ is the collection of attention weight $a_{x_{i},e_{k}}$. 

The adopted attention mechanism $\operatorname{Attention}(\cdot )$ can be designed similarly to GATs \cite{GATs}. We first concatenate the node embedding vector and the hyperedge embedding vector and then employ a weight vector $\alpha_{X} \in \mathbb{R}^{2d \times 1}$ to map it to a scalar value, which can be formulated as
\begin{equation}
\label{eq16}
\begin{array}{l}
    \mathit{DA}_{x_{i},e_{k}}=\quad \frac{\exp \left(\operatorname{LeakyReLU}\left(\alpha_{X} ^{\top }\left(W x_{i} \parallel  W e_{k}\right)\right)+\tilde{\rho}_{x_{i}}\right)}
    {\sum_{x_{j} \in \mathcal{N}\left(e_{k}\right)} \exp \left(\operatorname{LeakyReLU}\left(\alpha_{X}^{\top}\left(W x_{j} \parallel  W e_{k}\right)\right)+\tilde{\rho}_{x_{j}}\right)},
\end{array}
\end{equation}
where $\mathcal{N}\left(e_{k}\right)$ denotes the set of vertices connected by the hyperedge $e_{k}$. $\operatorname{LeakyReLU}(\cdot)$ is an activation function. And $\parallel$ represents the concatenation operation. 

Then we can obtain the density-aware attention matrix $\mathit{DA}_{X}\in \mathbb{R}^{n \times m}$, of which each element is $\mathit{DA}_{x_{i},e_{k}}\in \left [0,1 \right ] $. At last, we utilize this density-aware attention matrix to perform feature aggregation, which is formulated as follows:
\begin{equation}
\label{eq17}
    \tilde{E}  = \sigma (\mathit{DA}_{X}^{\top }WX),
\end{equation}
where $\sigma(\cdot)$ is an activation function, which can be $\operatorname{ELU}(\cdot)$ in implementation.

\subsubsection{Density-aware attention hyperedge aggregation} 
Density-aware attention hyperedge aggregation module integrates the density information of each hyperedge as a part of the attention and then aggregates the connected hyperedge to enhance the node embedding. Similarly, we also define a density rule for each hyperedge. The density of each hyperedge can be defined as the sum of the density of all nodes connected by this hyperedge, which is formulated as
\begin{equation}
\label{eq18}
    \rho_{e_{k}}=\sum_{x_{j} \in \mathcal{N}\left(e_{k}\right)} \rho_{x_{j}}.
\end{equation}

Intuitively, if a hyperedge has a higher density, it would be located in a node-dense area. %Thus, when performing hyperedge feature aggregation, we need to pay more attention to it.
Accordingly, in the density-aware attention hyperedge aggregation module, we calculate the density-aware attention weights of each hyperedge with respect to each node connected by this hyperedge. We employed an attention mechanism similar to the above density-aware attention vertex aggregation module, that is,
\begin{equation}
\label{eq19}
\begin{array}{l}
\mathit{DA}_{e_{k}, x_{i}}=\quad \frac{\exp \left(\operatorname{LeakyReLU}\left(\alpha_{E}^{\top }\left(W e_{k} \parallel W x_{i} \right)\right)+\tilde{\rho}_{e_{k}}\right)}
{\sum_{e_{j} \in \mathcal{N}\left(x_{i}\right)} \exp \left(\operatorname{LeakyReLU}\left(\alpha_{E}^{\top}\left(W e_{j} \parallel W x_{i} \right)\right)+\tilde{\rho}_{e_{\mathbf{j} }}\right)},
\end{array}
\end{equation}
where $\mathcal{N}\left(x_{i}\right)$ represents the set of hyperedges connecting to vertex $x_{i}$. $\alpha_{E} \in \mathbb{R}^{2d \times 1}$ is a weight vector to be trained. And $\tilde{\rho}_{e_{\mathbf{k} }}$ is the normalized density.

Afterward, we can obtain the  density-aware attention matrix $\mathit{DA}_{E}\in \mathbb{R}^{m \times n}$, which can be utilized to aggregate hyperedge features and update the node embedding by:
\begin{equation}
\label{eq20}
    \tilde{X}  = \sigma (\mathit{DA}_{E}^{\top } \tilde{E}).
\end{equation}

We combine the two modules described above to form a density-aware hypergraph attention layer shown as follows:
\begin{equation}
\tilde{X}=\operatorname{ELU}\left(\mathit{DA}_{E}^{\top } \ \operatorname{ELU}\left(\mathit{DA}_{X}^{\top } W X \right)\right).
\end{equation}

In each density-aware hypergraph attention layer, we first pay a density-aware attention weight to each node and gather the node features to enhance hyperedge features. Then we assign a density-aware attention weight to each hyperedge and aggregate the connected hyperedge features to generate new vertex features. By using this node-hyperedge-node feature transform mechanism, we can efficiently explore high-order semantic correlation among data.

This study chooses the multi-head attention mechanism \cite{GATs} to enhance the density-aware hypergraph attention layer. The output feature representation of this layer is obtained by concatenating the output features of each head, that is,
\begin{equation}
\tilde{X}= \parallel_{t=1}^{T} \operatorname{ELU}\big(\mathit{DA}_{E}^{\top } \ \operatorname{ELU}(\mathit{DA}_{X}^{\top } W X)\big),
\end{equation}
where $\parallel_{t=1}^{T}$ denotes the concatenation operation, and $T$ is the number of attention heads. The final output of the DualHGNN is a low-dimensional node embedding and the class prediction $Z \in \mathbb{R}^{n \times c}$ can be obtained by performing a $\operatorname{softmax}(\cdot)$.

The cross-entropy loss is adopted as the optimization function:
\begin{equation}
\label{eq23}
\mathcal{L}_{CE}=-\sum_{i \in L} \sum_{j=1}^{c} Y_{i j} \ln Z_{i j},
\end{equation}
where $L$ is the set of labeled samples.

Accordingly, we will jointly optimize the multi-view hypergraph learning network and the density-aware hypergraph attention network by linearly combining the hypergraph learning loss and the cross-entropy loss, which is shown as follows:
\begin{equation}
\label{eq24}
\mathcal{L}=\mathcal{L}_{HGL} + \lambda \mathcal{L}_{CE},
\end{equation}
where $\lambda$ is a trade-off parameter.
Overall, the entire algorithm of DualHGNN is summarized in Algorithm~\ref{DualGHNN-Alg}.

\RestyleAlgo{ruled}
\begin{algorithm}[!ht]
% \small
  \caption{The algorithm of DualHGNN.}
  \label{DualGHNN-Alg}
    \KwIn{Node feature matrix  $X_{0}\in \mathbb{R}^{n \times d}$, initial hypergraph incidence matrix $H_{0}\in \mathbb{R}^{n \times m}$.} 
    \KwSty{Initialize} $P$, $W$, $\alpha_{X}$, $\alpha_{E}$\\
    
    \While{not converges}{
    \For{each view of hypergraph learning network}{
    Calculate $S$ using Eq.(\ref{eq2}). \\
    Sparse sampling to obtain $\tilde{S}$ using Eq.(\ref{eq3}). \\
    Construct hyperedges based on $\tilde{S}$. \\
    }
    Calculate the learned $H$ using Eq.(\ref{eq5}). \\
    Calculate $\tilde{H}$ using Eq.(\ref{eq8}). \\
    
    Perform hypergraph message propagation to update $X$ and $E$ according to Eq.(\ref{eq10}) and Eq.(\ref{eq11}). \\
    \For{each density-aware hypergraph attention layer}{
        Calculate node density $\rho_{X}$ using Eq.(\ref{eq12}). \\
        Calculate $\mathit{DA}_{X}$ using Eq.(\ref{eq16}). \\
        Perform attention vertex aggregation according to Eq.(\ref{eq17}). \\
    
        Calculate hyperedge density $\rho_{E}$ using Eq.(\ref{eq18}). \\
        Calculate $\mathit{DA}_{E}$ using Eq.(\ref{eq19}). \\
        Perform attention hyperedge aggregation according to Eq.(\ref{eq20}). \\
    }
    Calculate $\mathcal{L}_{HGL}$ using Eq.(\ref{eq9}). \\
    Calculate $\mathcal{L}_{CE}$ using Eq.(\ref{eq23}). \\
    Calculate $\mathcal{L}$ using Eq.(\ref{eq24}). \\
    Update parameters by performing back-propagation.
    }
\end{algorithm}

\section{Experiments}
\label{Experiments}

\subsection{Datasets}
We evaluate the effectiveness of our method on three widely-used image datasets: Scene15 \cite{DBLP:conf/cvpr/LazebnikSP06}, CIFAR-10 \cite{krizhevsky2009learning}, and MNIST \cite{lecun1998gradient}. Each dataset is used in a semi-supervised learning setup where only a small part of the data samples are labeled. More details of these datasets and their usage in our experiments are introduced as follows, which are also summarized in Table~\ref{datasets-table}.

\textbf{Scene15:} This dataset contains 4,485 RGB images coming from 15 scene categories, and each category contains 200 to 600 samples. In our experiments, we used all 4,485 samples to evaluate our method. For each image, we use the 3,000-dimension features provided in the previous work \cite{DBLP:journals/pami/JiangLD13}.

\textbf{CIFAR-10:} This dataset consists of 10 types of natural images. In our experiments, we use 10,000 images from the independent test set to evaluate our method. To represent each image, we used the same 13-layer CNN networks as in \cite{DBLP:conf/cvpr/IscenTAC19} to extract the features.

\textbf{MNIST:} It contains 10 classes of images of hand-written digits. We randomly selected 1,000 images for each class and obtained 10,000 images at to conduct our experiments. Similar to the prior work \cite{DGL,GLCN}, we use 784-dimension feature vectors converted from grayscale to represent each sample.

% For the Scene15 dataset, we randomly selected 250, 500,  750, and 1,000 images as labeled samples. For unlabeled samples, we randomly pick 500 images used for validation, and the remaining 3735, 3485, 3235, and 2985 images are used as test samples.
% For CIFAR-10 and  MNIST datasets, we randomly select 50, 100, 200, 300, and 400 images per class and get 500, 1,000, 2000, 3000, and 4000 images as labeled samples. For unlabeled samples, we pick 100 images at random per class and get 1,000 images used for validation. The remaining 8500, 8000, 7000, 6000, and 5000 images are used as test samples. 

\begin{table*}[ht]
\centering
\caption{Datasets statistics and the extracted features in experiments.}
\label{datasets-table}
\begin{tabular}{l|cccccc}
\hline
Dataset & Samples & Training Samples & Validating Samples & Testing Samples & Classes & Features \\
\hline
Scene15  &   4,485 &  250 / 500 / 750 / 1,000 & 500 & 3,735 / 3,485 / 3,235 / 2,985 & 15 & 3,000 \\
\hline
CIFAR-10  &  10,000 & 500 / 1,000 / 2,000 / 3,000 / 4,000 & 1,000 & 8,500 / 8,000 / 7,000 / 6,000 / 5,000 & 10 & 128 \\
\hline
MNIST  &  10,000  &  500 / 1,000 / 2,000 / 3,000 / 4,000 & 1,000 & 8,500 / 8,000 / 7,000 / 6,000 / 5,000 & 10 & 784 \\
\hline
\end{tabular}
\end{table*}

\begin{table*}[!htbp]
\centering
\caption{Classification accuracy ($\%$) on Scene15, CIFAR-10 and MNIST datasets with different number of labeled samples.}
\label{comparison-table}
\begin{tabular}{l|lccccc}
\hline 
\hline
Datasets & No. of labels & 250 & 500 & 750 & 1,000 \\ 
\hline
\multirow{12}{*}{Scene15} & GCNs \cite{GCNs}    &  89.96±1.17  &  94.02±1.04  &  94.75±0.96  &  95.86±0.64  \\
~ & GATs \cite{GATs}   &  97.06±0.44  &  98.01±0.28  &  98.26±0.25  &  98.32±0.18 \\
~ & GraphSAGE \cite{GraphSAGE}   &	95.95±0.87	&  97.74±0.32	&  97.98±0.24	&  98.26±0.26  \\
~ & APPNP \cite{APPNP}	 &	96.44±0.73	 &	97.39±0.32	 &	97.59±0.27	 &	97.89±0.25 \\
~ & HGNN \cite{HGNN}   &  90.49±1.47   &  94.11±0.41  &  94.52±0.47  & 95.58±0.52 \\
~ & DHGNN \cite{DHGNN} &	94.29±0.51	&   95.14±0.26	&   95.42±0.35	&   95.58±0.29 \\
~ & SGC	\cite{SGC} &	95.34±0.60	&	96.27±0.56	&	96.41±0.52	&	96.89±0.44\\
~ & DropEdge \cite{DropEdge}	&	85.04±2.09	&	91.55±0.63	&	93.16±0.53	&	94.16±0.62\\
~ & GCNII \cite{GCNII}	&	96.03±1.58	&	96.64±1.90	&	97.10±1.33	&	97.30±1.60\\
~ & GRAND \cite{GRAND}	& 	88.46±0.81	& 	90.41±0.85	&  	91.12±0.67	&  	91.69±0.56 \\
~ & ElasticGNN \cite{ElasticGNN}	&   96.46±0.08	&   96.54±0.16	&   97.09±0.12  &   	97.25±0.09 \\
~ & DualHGNN (ours)        & \textbf{98.55±0.09}   &  \textbf{98.71±0.06}  & \textbf{98.79±0.07}   &  \textbf{98.85±0.10}  \\
\hline
\hline
Datasets & No. of labels & 500 & 1,000 & 2,000 & 3,000 & 4,000 \\ 
\hline
\multirow{12}{*}{CIFAR-10} & GCNs \cite{GCNs}    & 91.48±0.25  &  91.70±0.12  &  92.55±0.13  &  92.64±0.14  & 92.98±0.16  \\
~ & GATs \cite{GATs}   & 93.80±0.13  &  93.59±0.42  &  93.90±0.13  &  93.97±0.09  & 93.80±0.04  \\
~ & GraphSAGE \cite{GraphSAGE}	 &  92.73±0.10	 &  92.49±0.21	 &  92.08±0.11 &  	92.04±0.22	 &  92.15±0.17 \\
~ & APPNP \cite{APPNP}	&  92.46±0.56	&  92.52±0.32	&  92.69±0.28	&  92.97±0.33	&  92.80±0.37\\
~ & HGNN \cite{HGNN}   & 90.97±0.41  & 91.26±0.19   &  91.35±0.30  &  91.68±0.11  & 91.85±0.33  \\
~ & DHGNN \cite{DHGNN}  &  	93.95±0.13   &  	93.77±0.26   &  	93.88±0.18   &  	93.95±0.13	   &  93.76±0.17  \\
~ & SGC	\cite{SGC}   & 90.64±0.25	&  92.24±0.17	&  93.32±0.32	&  94.19±0.08	&  94.15±0.07\\
~ & DropEdge \cite{DropEdge}	&  91.99±0.46	&  92.42±0.34	&  92.88±0.24	&  93.04±0.32	&  93.25±0.31\\
~ & GCNII \cite{GCNII}	&  92.99±0.26	&  93.10±0.23	&  93.06±0.40	&  93.24±0.33	&  93.13±0.33\\
~ & GRAND \cite{GRAND}	&  93.57±0.14	&  93.74±0.19	&  93.88±0.19	&  93.79±0.15	&  93.87±0.18\\
~ & ElasticGNN	\cite{ElasticGNN}   &  93.81±0.24	   &  94.03±0.09	   &  93.92±0.25   &  	94.15±0.07   &  	94.03±0.09  \\
~ & DualHGNN (ours)      &  \textbf{93.99±0.14}   & \textbf{94.08±0.07}   &  \textbf{94.10±0.17} &  \textbf{94.24±0.16}   &  \textbf{94.19±0.33}  \\
\hline 
\hline
Datasets &  No. of labels & 500 & 1,000 & 2,000 & 3,000 & 4,000 \\ 
\hline
\multirow{12}{*}{MNIST} & GCNs \cite{GCNs}    &  90.37±0.32  &  90.42±0.39  &  90.28±0.34  &  90.30±0.38  &  90.20±0.32  \\
~ & GATs \cite{GATs}   &  91.40±0.14  &  92.44±0.12  &  92.99±0.20  &  93.05±0.18  & 93.41±0.28 \\
~ & GraphSAGE \cite{GraphSAGE}	&  89.74±0.92	&  90.72±0.60	&  91.88±0.25   &  	92.27±0.60	&  92.31±0.69 \\
~ & APPNP \cite{APPNP}	&  86.17±1.14	&  86.21±1.28	&  86.70±1.16	&  87.75±1.16	&  87.67±0.88\\
~ & HGNN \cite{HGNN}   &  88.70±0.46   &  90.26±0.53  &  91.34±0.45  &  92.25±0.32  & 92.41±0.34 \\
~ & DHGNN \cite{DHGNN}	&  86.68±2.38	&  88.64±0.91	&  89.40±1.46 &  	89.17±1.38	&  89.78±1.36  \\
~ & SGC \cite{SGC}	&  90.11±1.17	&  91.83±0.65	&  93.21±0.32	&  94.21±0.32	&  94.44±0.54\\
~ & DropEdge \cite{DropEdge}  &  88.98±0.85	 &  90.96±0.27	&   92.21±0.50	&  93.09±0.43	&  93.44±0.52 \\
~ & GCNII \cite{GCNII}	&  86.93±1.51	&  87.67±1.18	&  87.70±1.23	&  88.38±1.49	&  88.73±1.23 \\
~ & GRAND   \cite{GRAND}	&  84.66±0.91	&  86.80±0.88	&  88.41±0.90	&  89.33±0.82	&  89.91±1.15\\
~ & ElasticGNN \cite{ElasticGNN}	&  93.25±0.17	&  93.77±0.14	&  \textbf{95.09±0.06}  &  	94.88±0.21	&  \textbf{95.71±0.04}  \\
~ & DualHGNN (ours)        & \textbf{93.57±0.29}   &  \textbf{94.43±0.15}  & 94.56±0.20   &  \textbf{94.89±0.15}  &  94.70±0.07  \\
\hline 
\hline
% Datasets & No. of labels & 500 & 1,000 & 2000 & 3000 & 4000 \\ 
% \hline
% \multirow{12}{*}{SVHN} & GCNs \cite{GCNs}    & 95.52±0.12  &  95.17±0.20  & 95.54±0.11  &  95.70±0.11    & 95.61±0.16  \\
% ~ & GATs \cite{GATs}   & 96.15±0.09  &  96.23±0.10  &  96.33±0.04 &   96.38±0.13    &   96.14±0.02  \\
% ~ & GraphSAGE \cite{GraphSAGE}	&  96.09±0.11	&  96.18±0.09	&  96.12±0.29   &  	95.96±0.08	&  95.97±0.15  \\
% ~ & APPNP \cite{APPNP}	&  95.58±0.49	&  95.72±0.39	&  95.91±0.21	&  95.96±0.28	&  96.04±0.29\\
% ~ & HGNN \cite{HGNN}   & 94.20±0.58 & 94.61±0.53  & 94.73±0.49  & 94.73±0.33  &  94.61±0.35  \\
% ~ & DHGNN \cite{DHGNN}  & 	\textbf{96.61±0.15}   & 	96.68±0.18   & 	96.66±0.16   & 	96.67±0.20   & 	96.64±0.22 \\
% ~ & SGC	\cite{SGC}   & 94.94±0.68	& 95.77±0.39	& 96.21±0.17	& 96.60±0.08	& 96.54±0.17\\
% ~ & DropEdge \cite{DropEdge}	& 95.06±0.51	& 95.27±0.4	& 95.86±0.31	& 95.94±0.31	& 96.01±0.34 \\
% ~ & GCNII \cite{GCNII}	& 96.16±0.33	&  96.25±0.21	&  96.35±0.2	 &  96.3±0.27	&  96.38±0.21 \\
% ~ & GRAND   \cite{GRAND}	& 	96.33±0.26	& 	96.4±0.29	& 	96.35±0.21	& 	96.44±0.16	& 	96.53±0.28\\
% ~ & ElasticGNN \cite{ElasticGNN}	   & 96.48±0.04	   & 96.49±0.10	   & 96.65±0.04   & 	96.64±0.07	   & \textbf{97.21±0.04} \\
% ~ & DualHGNN (ours)       & 96.55±0.05  & \textbf{96.69±0.12}  &  \textbf{96.67±0.11}  &  \textbf{96.70±0.17}  &  96.71±0.14 \\
% \hline 
% \hline
\end{tabular}
\end{table*}

\subsection{Experiment Setup}
\label{Experimental_setting}

For the architecture of DualHGNN, we adopt a multi-view hypergraph learning network with two views to learn a hypergraph and employ a two-layer density-aware hypergraph attention network for hypergraph representation learning where the first layer uses a multi-head attention mechanism with two heads. 
The similarity measure functions we adopt in the multi-view hypergraph learning network are cosine similarity and inner product.
The output low dimension of $P$ is set to 70 for the Scene15 and CIFAR-10 datasets and to 128 for the MNIST dataset.
We introduce $L2$-normalization into each view of the hypergraph learning network and each density-aware hypergraph attention layer.
The number of units in the density-aware hypergraph attention hidden layer is set to $64$. We employ Xavier algorithm \cite{DBLP:journals/jmlr/GlorotB10} for the initialization of $P$, $W$, $\alpha_{X}$ and $\alpha_{E}$.
We adopt Adam optimizer \cite{DBLP:journals/corr/KingmaB14} with a learning rate of 0.2, 0.01, and 0.002 for Scene15, CIFAR-10, and MNIST datasets, respectively,  and the learning rate decays to half after every 100 epochs. We train DualHGNN for a maximum of 2,000 epochs and stop training if the validation loss does not decrease for 100 consecutive epochs.

\subsection{Performance}

\textbf{Baselines}: We compare the proposed DualHGNN with representative graph-based semi-supervised node classification methods, including GCNs \cite{GCNs}, GATs \cite{GATs}, GraphSAGE \cite{GraphSAGE}, APPNP \cite{APPNP}, HGNN \cite{HGNN}, DHGNN \cite{DHGNN}, SGC \cite{SGC}, DropEdge \cite{DropEdge}, GCNII\cite{GCNII}, GRAND \cite{GRAND} and ElasticGNN \cite{ElasticGNN}. For a fair comparison, we construct a k-nearest neighbor graph for all the methods, and the value of $k$ is set to 15. We retrain all the baseline methods, and all the reported results are averaged over 10 runs.

\textbf{Results}: Table~\ref{comparison-table} summarizes the classification accuracy comparison results on three datasets. The best results are highlighted. From these results, we can make a few observations as follows. 
First, in Scene15 and CIFAR-10 datasets, our DualHGNN significantly outperforms all the baseline approaches. Compared with the state-of-the-art method ElasticGNN, our DualHGNN achieved at least 1.6\% and at most 2.17\% improvement on the Scene15 dataset. This may be because the data points of each category in the Scene15 dataset are imbalanced, and the baseline methods suffer performance in imbalanced data, while our DualHGNN still maintains excellent performance. DualHGNN is slightly better than ElasticGNN on CIFAR-10 dataset, while it achieves a lower standard deviation in most cases. These clearly prove the strong performance of DualHGNN on graph-based semi-supervised node classification. 
In the MNIST dataset, DualHGNN outperforms most baseline methods. Compared with ElasticGNN, DualHGNN also obtains competitive performance, especially in the case of a few labeled samples, such as less than 1,000 labels. This also proves the advantage of DualHGNN in the case of fewer labels. 
Moreover, in all three datasets, DualHGNN outperforms the baseline method GCNs and GATs by significant margins. Compared with GATs, DualHGNN receives an improvement at most of 1.49\%, 0.49\%, and 2.17\% on the Scene15, CIFAR-10, and MNIST datasets, respectively. DualHGNN significantly outperforms the hypergraph neural networks baseline HGNN with the least margins of 3.27\%, 2.34\%, and 2.29\% on the Scene15, CIFAR-10, and MNIST datasets, respectively, which straightforwardly indicates the higher predictive accuracy on semi-supervised node classification of DualHGNN by performing multi-view hypergraph learning and density-aware attention neighborhood aggregation.

\subsection{Ablation Study}

\subsubsection{Effectiveness of multi-view hypergraph learning network} 
\begin{figure}
\begin{minipage}[h]{1.0\linewidth}
    \centering
    \subfigure[]{
    \includegraphics[width=.48\linewidth]{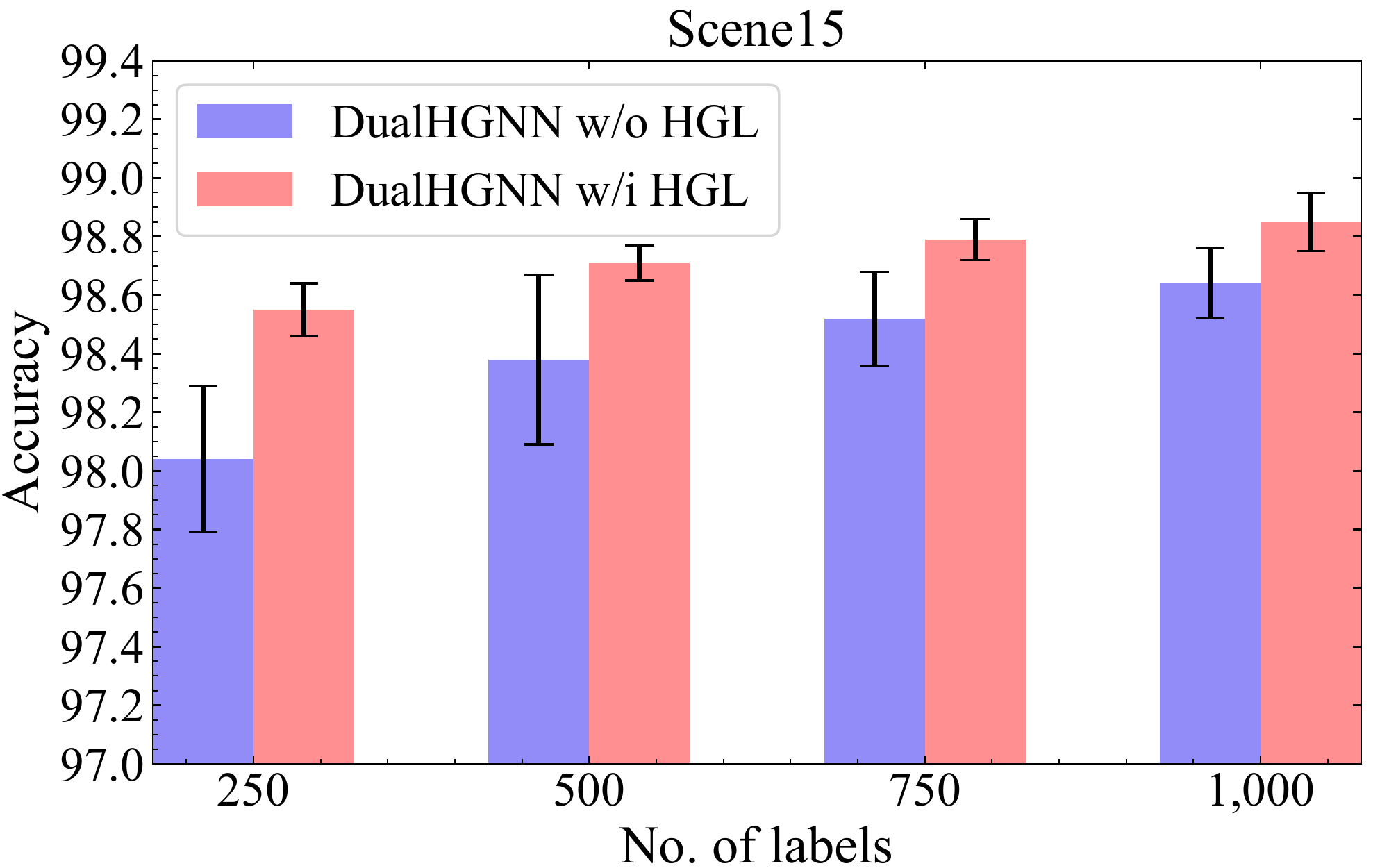}
    }%
    \subfigure[]{
    \includegraphics[width=.48\linewidth]{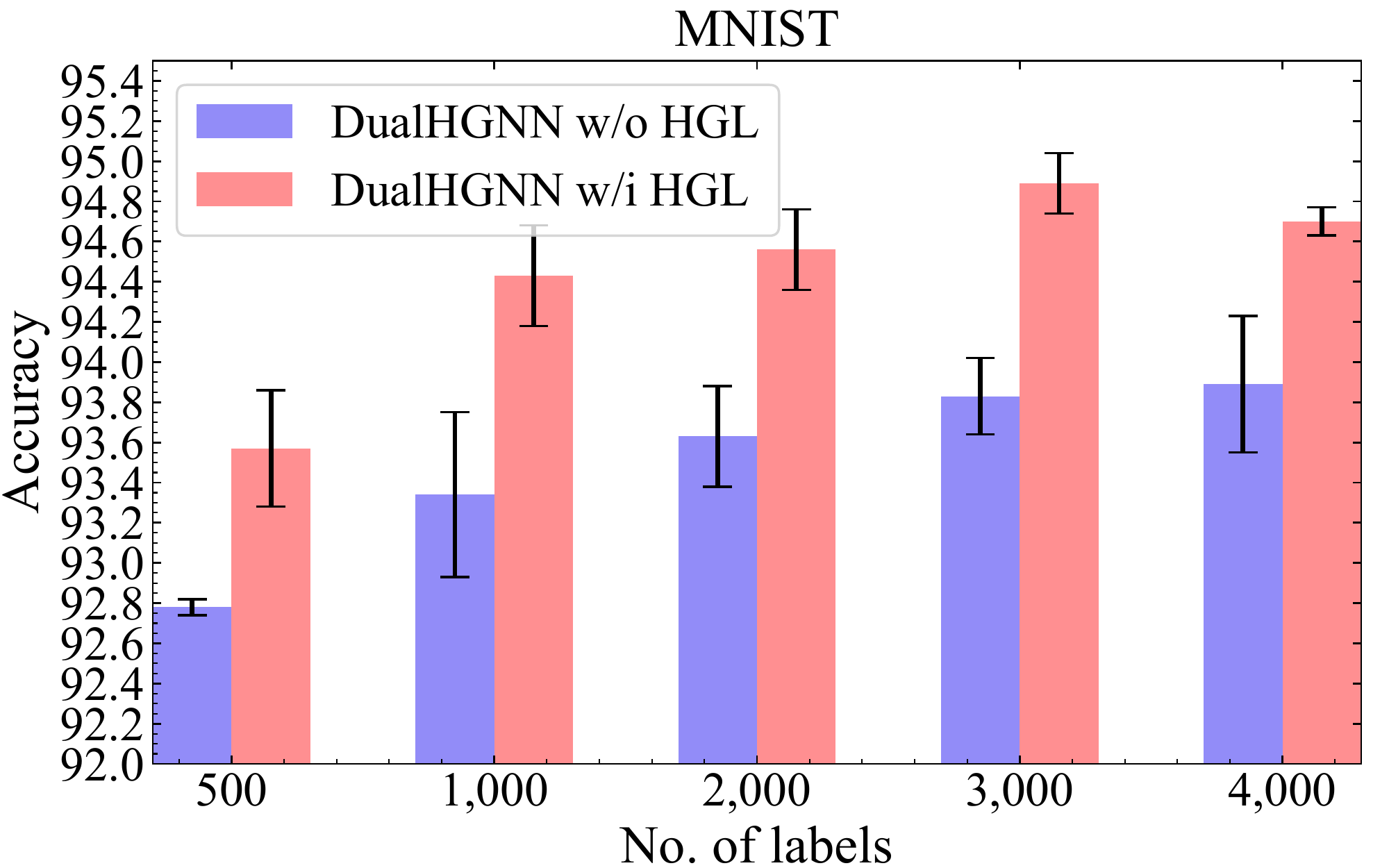}
    }
\end{minipage}
\caption{The average accuracy of DualHGNN within and without HGL on (a) the Scene15 dataset and (b) the MNIST dataset.}
\label{ablation-study-HGL}
\end{figure}

To verify the effectiveness of the multi-view hypergraph learning network, we conducted an ablation study on the multi-view hypergraph learning network. We removed the multi-view hypergraph learning network from the proposed DualHGNN and denoted it as \emph{DualHGNN w/o HGL}, which performs hypergraph representation learning only based on an original $k$-NN hypergraph. The proposed version is denoted as \emph{DualHGNN w/i HGL}. The ablation experiments are conducted on all three datasets, and for ease of presentation, we only show the results on Scene15 and MNIST datasets in Figure~\ref{ablation-study-HGL}.
From these results, we can clearly observe that employing the multi-view hypergraph learning network to learn the hypergraph structure can achieve higher classification accuracy than only using the original $k$-NN hypergraph and receives an improvement at most of 0.51\% and 1.09\% on the Scene15 and MNIST datasets, respectively. This further demonstrates the effectiveness of the proposed multi-view hypergraph learning network.

\subsubsection{Effectiveness of density-aware attention mechanism} 
\begin{figure}
\begin{minipage}[h]{1.0\linewidth}
    \centering
    \subfigure[]{
    \includegraphics[width=.48\linewidth]{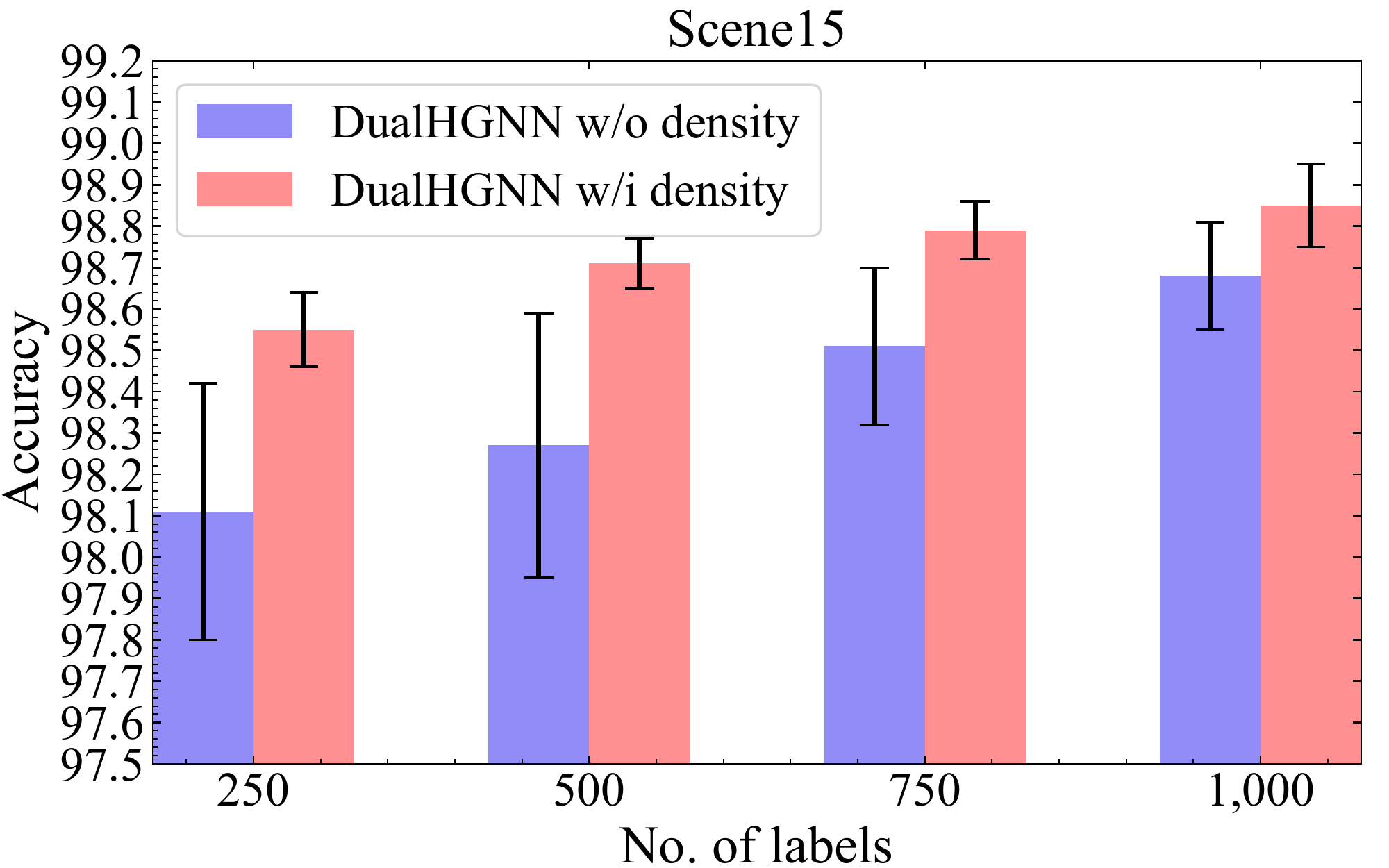}
    }%
    \subfigure[]{
    \includegraphics[width=.48\linewidth]{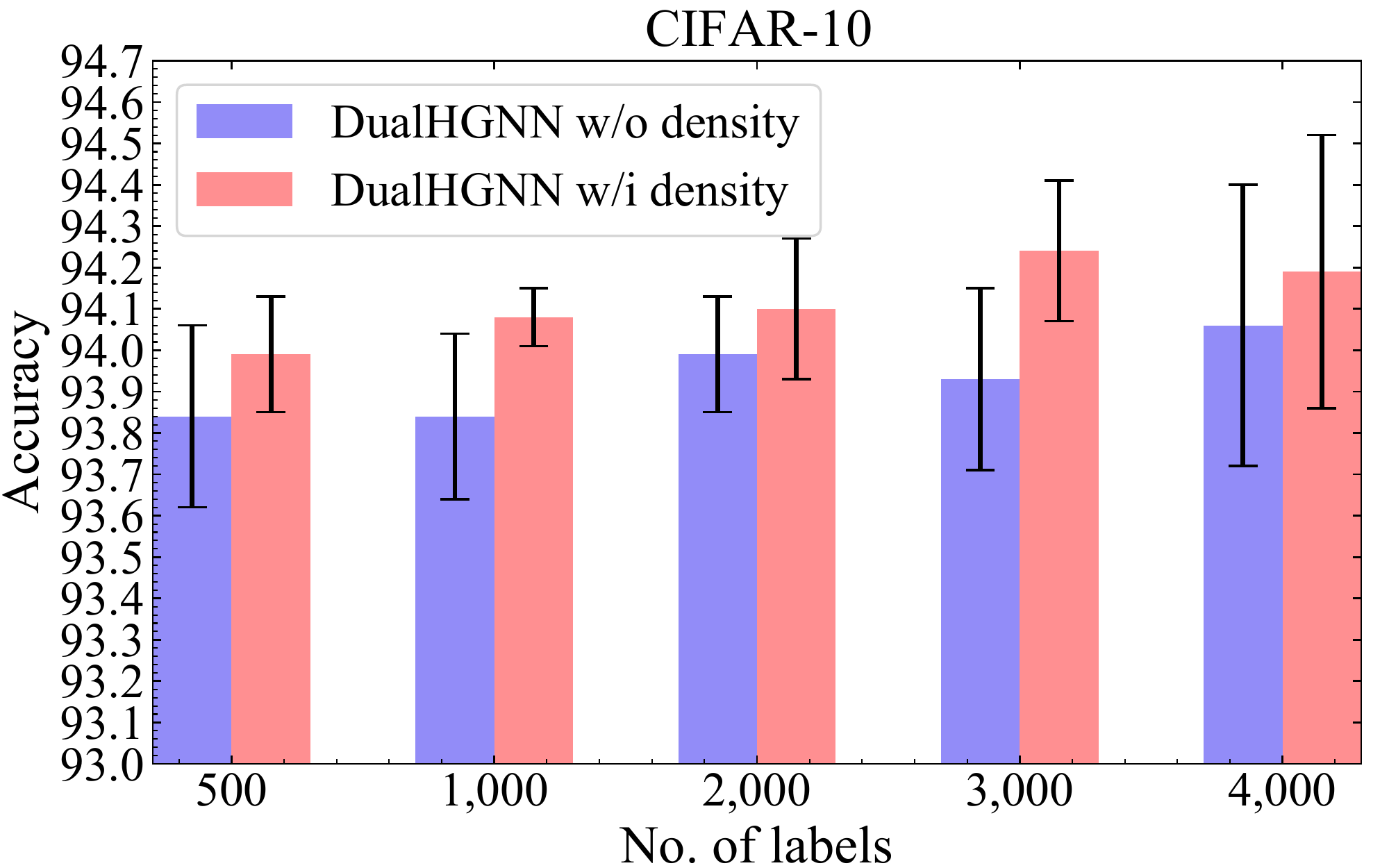}
    }%
\end{minipage}
\caption{The average accuracy of DualHGNN within and without density on (a) the Scene15 dataset and (b) the CIFAR-10 dataset.}
\label{ablation-study-density}
\end{figure}

We also conducted an ablation study on all three datasets to evaluate the effectiveness of the proposed density-aware attention mechanism. We remove the density information from the proposed DualHGNN and only keep the traditional attention mechanism, denoted as \emph{DualHGNN w/o density}. Correspondingly, the proposed version is denoted as \emph{DualHGNN w/i density}. Similarly, given the page limitation, we only show the results on Scene15 and CIFAR-10 datasets in Figure~\ref{ablation-study-density}. 
From these results, we can observe that integrating the density-aware attention mechanism can significantly improve the performance of hypergraph representation learning and achieve higher predictive accuracy. This directly demonstrates the effectiveness of the density-aware attention mechanism on graph-based semi-supervised node classification.

\begin{figure}
\begin{minipage}[h]{1.0\linewidth}
\centering
\subfigure[]{
\includegraphics[width=.48\linewidth]{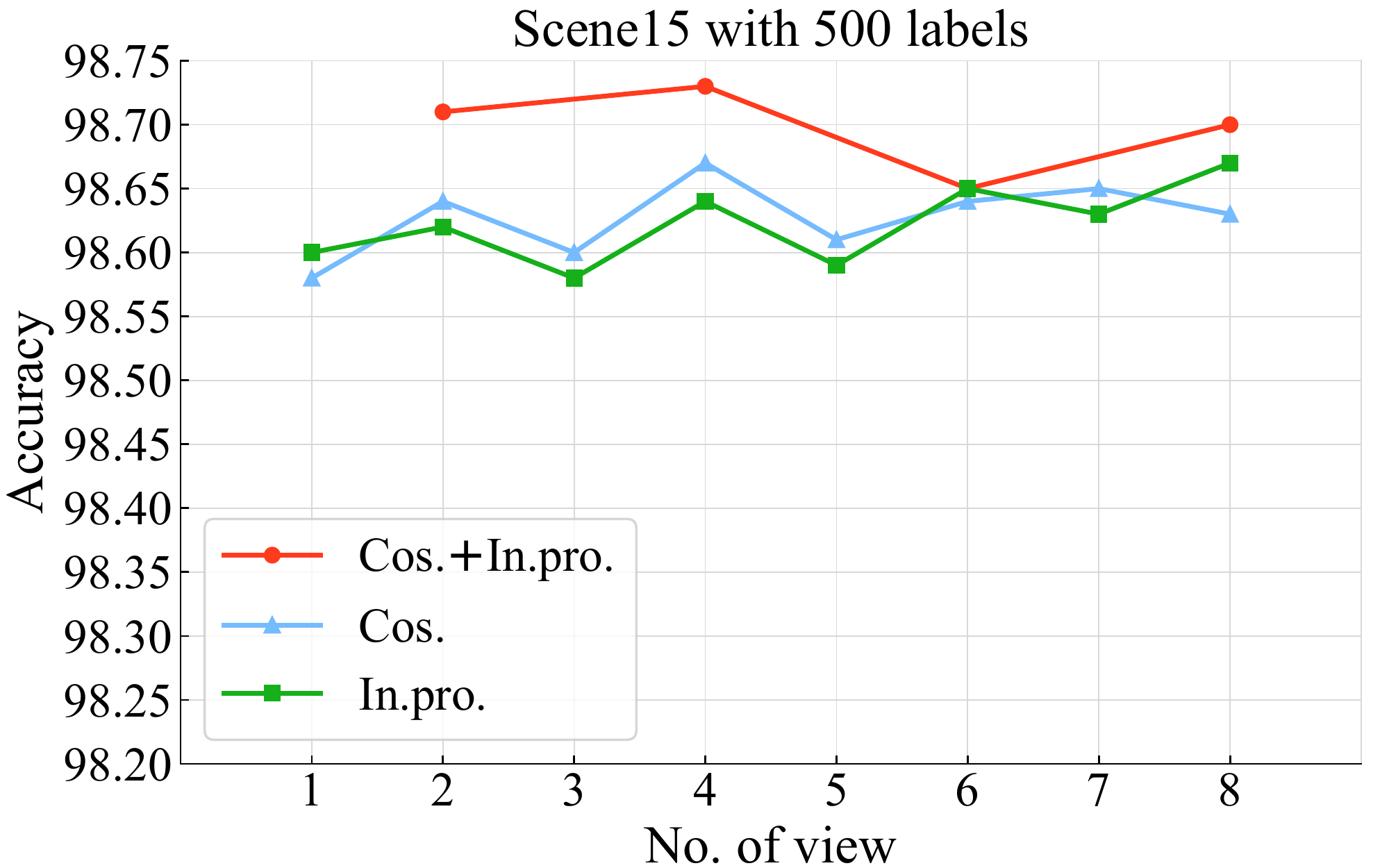}
}%
\subfigure[]{
\includegraphics[width=.48\linewidth]{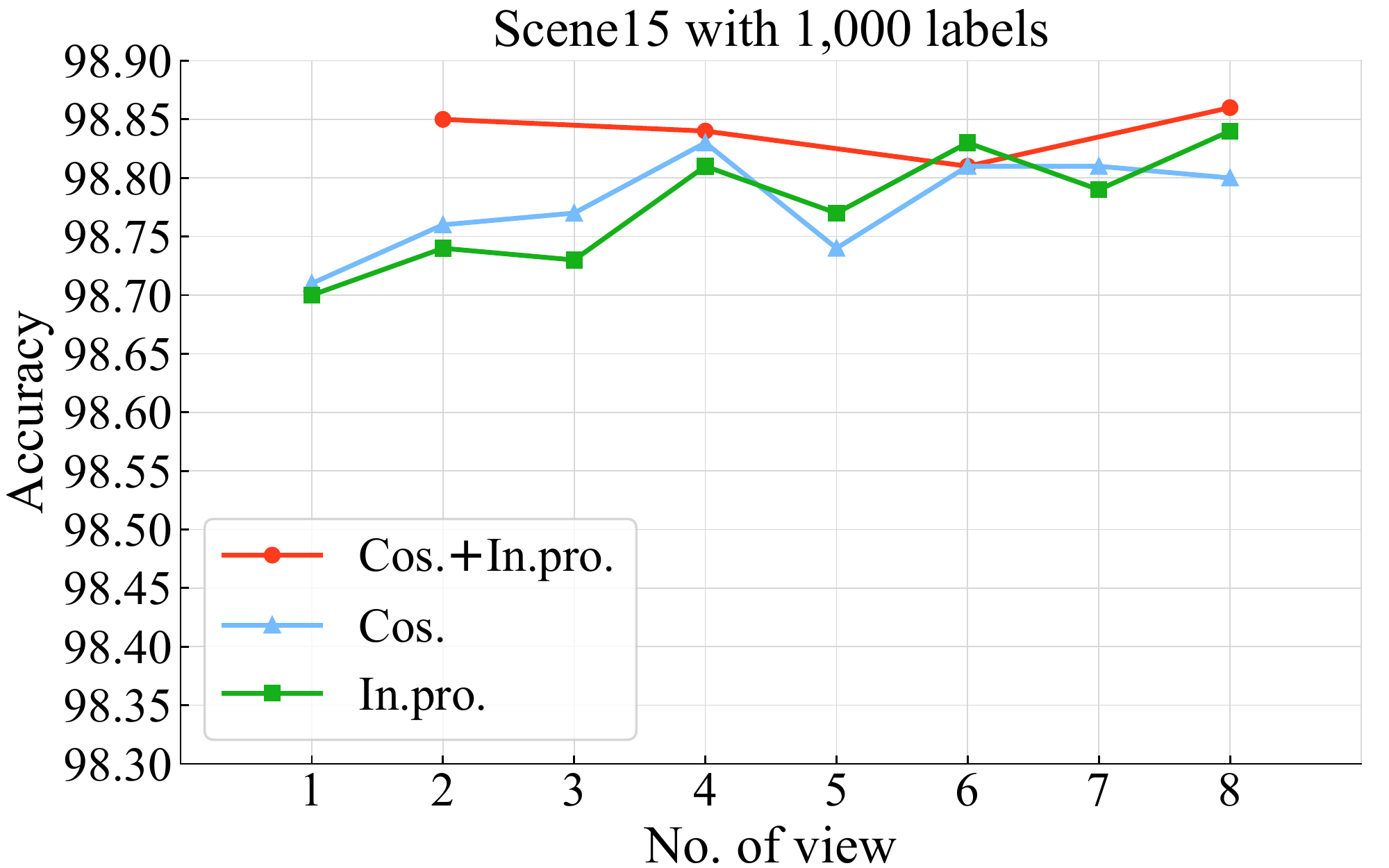}
}%
\end{minipage}
\caption{The average accuracy of DualHGNN with different multi-view hypergraph learning mechanisms and similarity measure functions on the Scene15 dataset with (a) 500 labeled samples and (b) 1,000 labeled samples.}
\label{Parameter_Analysis_HGL}
\end{figure}

\begin{figure}
\begin{minipage}[h]{1.0\linewidth}
\centering
\subfigure[]{
\includegraphics[width=.48\linewidth]{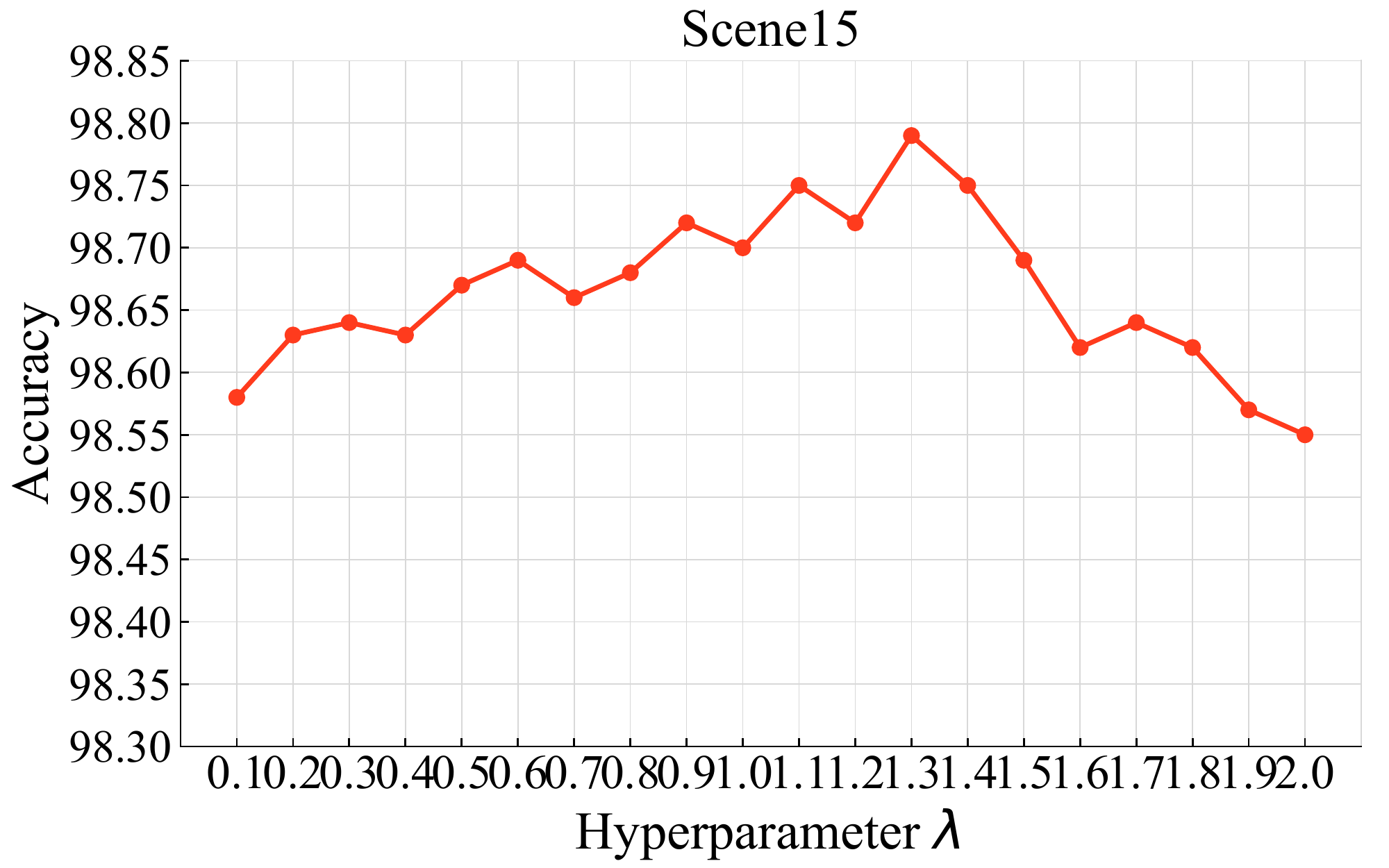}
}%
\subfigure[]{
\includegraphics[width=.48\linewidth]{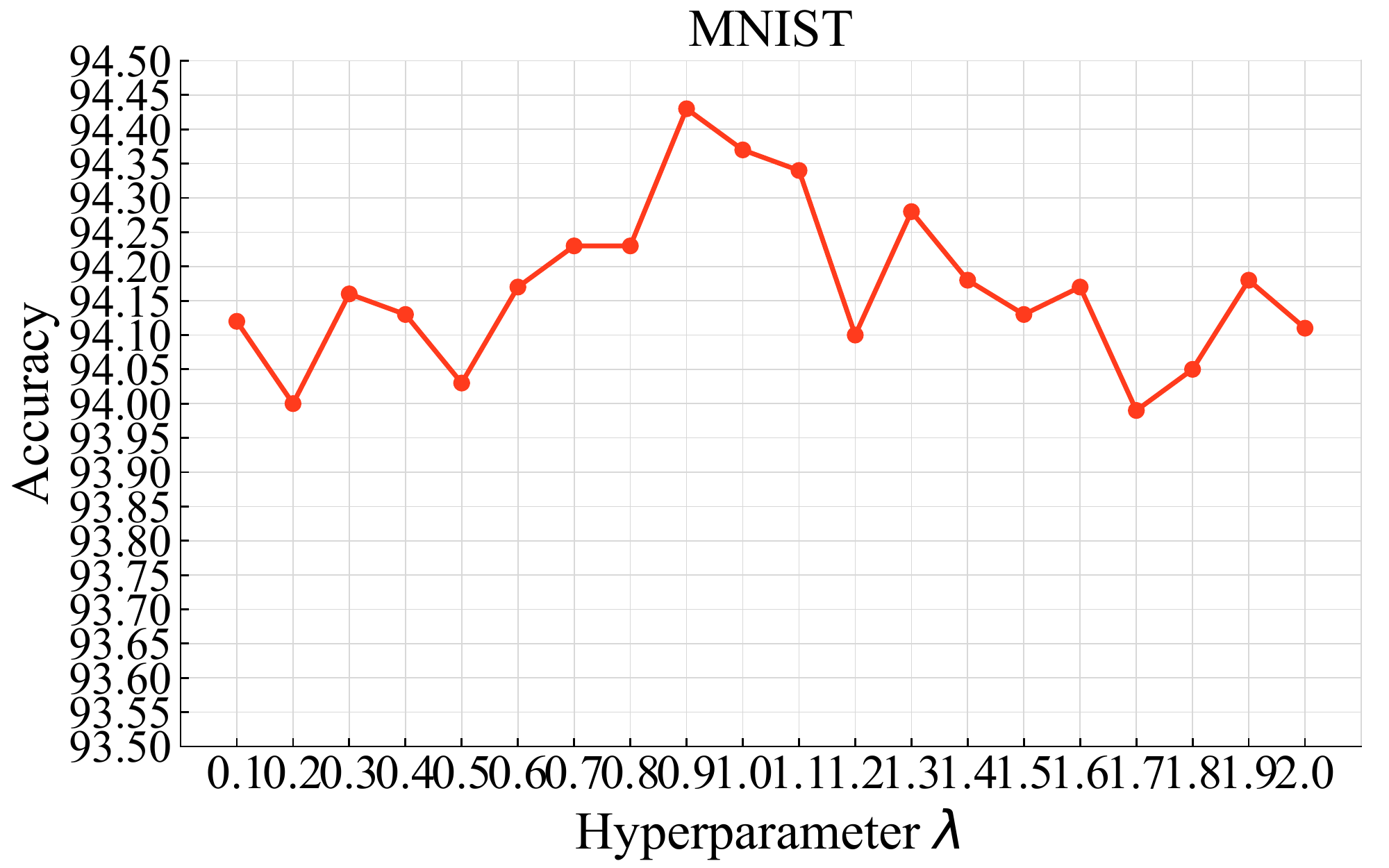}
}%
\end{minipage}
\caption{The parameter sensitivity to $\lambda$ on (a) the Scene15 dataset with 750 labeled samples and (b) the MNIST dataset with 1,000 labeled samples.}
\label{Parameter_Analysis_lamda}
\end{figure}

\subsection{Parameter Analysis}
\subsubsection{How the multi-view hypergraph learning network benefits the DualHGNN} Our multi-view hypergraph learning network is provided to learn a suitable hypergraph from multiple views with different similarity measure functions. To verify how the multi-view hypergraph learning network benefits the DualHGNN, we conduct some analysis experiments on the Scene15 dataset.
Specifically, we utilize both cosine similarity and inner product similarity measure functions at the same time to instantiate the multi-view hypergraph learning mechanism and denote it as ``\emph{Cos.+In.pro.}". The compared baselines adopt only one type of similarity measure function, and are denoted as ``\emph{Cos.}" or ``\emph{In-pro.}" accordingly.
We vary the number of views used in the hypergraph learning network from $1$ to $8$, and the experiment results are shown in Figure~\ref{Parameter_Analysis_HGL}.
It can be observed that combining cosine similarity and inner product similarity measure functions at the same time can significantly outperform those using only one of them, especially when fewer views are used in the hypergraph learning network. When using more views, adopting only one type of similarity measure function can also achieve satisfactory accuracy, but it also brings a large computational overhead. If both cosine similarity and inner product similarity measure functions are used at the same time, we can obtain satisfactory performance even if we only use two views.
In summary, utilizing different similarity measure functions in our multi-view hypergraph learning network to learn a hypergraph structure from multi-view can achieve higher classification performance and lower computing overhead, which once again demonstrates the effectiveness of the proposed multi-view hypergraph learning network.

\subsubsection{Effect of parameter $\lambda$} The proposed DualHGNN jointly optimizes the multi-view hypergraph learning network and the density-aware hypergraph attention network by linearly combining the two losses.
And the parameter $\lambda$ is a trade-off parameter of the hypergraph learning loss $\mathcal{L}_{HGL}$ and the cross-entropy loss $\mathcal{L}_{CE}$ in Eq.(\ref{eq24}).
We conduct a parameter analysis experiment to verify how different values of $\lambda$ influence the performance of DualHGNN.
For ease of presentation, we only show the results on Scene15 and MNIST datasets when setting $\lambda$ from $0.1$ to $2.0$ in Figure~\ref{Parameter_Analysis_lamda}.
It can be observed that choosing an appropriate value for $\lambda$ can increase the classification accuracy of DualHGNN to a certain extent, which is in line with our expectations of jointly optimizing the multi-view hypergraph learning network and the density-aware hypergraph attention network. However, setting $\lambda$ too large will also hurt the performance.
In our experiments, we set $\lambda = 1.3$, $1.1$ and $0.9$ on Scene15, CIFAR-10 and MNIST datasets, respectively.

\section{Conclusion}
\label{Conclusion}
In this paper, we propose a Dual Hypergraph Neural Network (DualHGNN), integrating both hypergraph structure learning and hypergraph representation learning simultaneously in a unified network architecture and performing joint optimization for semi-supervised node classification.
The DualHGNN first adopts a multi-view hypergraph learning network to learn a hypergraph structure from multi-view with different similarity measure functions.
Then DualHGNN employs a density-aware hypergraph attention network based on a density-aware attention mechanism to perform hypergraph representation learning.
We have conducted extensive experiments on three benchmark datasets and demonstrated the effectiveness of the DualHGNN on various semi-supervised node classification tasks.

Although our DualHGNN has achieved excellent performance, there are some limitations. For instance, the proposed multi-view learning mechanism introduces additional computing overhead. 
Besides, DualHGNN is only applied to node-level tasks, i.e., node classification.
% Besides, DualHGNN has conducted validity validation on multiple datasets, but they are all image-based hypergraph datasets.
For future work, on the one hand, we plan to apply our method to more and larger graph-based datasets and add the comparison in computing overhead at the same time to further validate its performance. On the other hand, we will attempt to extend it to graph-level tasks, such as graph classification.

\section*{Acknowledgment}
This work was supported by the National Natural Science Foundation of China (62276101) and the National Key R\&D Program of China (2019YFC1510400).

\bibliographystyle{IEEEtran}
\bibliography{DualHGNN}

\end{document}